\documentclass[accepted]{uai2021} 

\usepackage[american]{babel}

\usepackage{natbib} 
\usepackage{mathtools} 
\usepackage{booktabs} 
\usepackage{tikz} 

\usepackage{times}
\usepackage{epsfig}

\usepackage[utf8]{inputenc} 
\usepackage[T1]{fontenc}    
\usepackage{amsfonts}       
\usepackage{nicefrac}       
\usepackage{microtype}      
\usepackage{amsmath,amssymb}
\usepackage{bbm}
\usepackage{bm}
\usepackage{algorithm}
\usepackage{subfig}
\usepackage[font=small,labelfont=bf]{caption}

\usepackage{todonotes} 
\usepackage{graphics}
\usepackage{xcolor}
\usepackage{xspace}

\usepackage{algpseudocode}
\allowdisplaybreaks
\usepackage{amsthm}

\newcommand*{\eg}{e.g.\@\xspace}
\newtheorem{theorem}{Theorem}[section]

\newtheorem{lemma}[theorem]{Lemma}

\title{Unsupervised Program Synthesis for Images By Sampling Without Replacement}

\author{Chenghui Zhou} 
\author{Chun-Liang Li}
\author{Barnab\'{a}s P\'{o}czos}
\affil{%
    Machine Learning Department\\
    Carnegie Mellon University\\
    Pittsburgh, Pennsylvania, USA
    \texttt{\{chenghuz,chunlial,bapoczos\}@cs.cmu.edu}
    
}

\begin{document}
\maketitle


\begin{abstract}
  Program synthesis has emerged as a successful approach to the image parsing task. Most prior works rely on a two-step scheme involving supervised pretraining of a Seq2Seq model with synthetic programs followed by reinforcement learning (RL) for fine-tuning with real reference images. Fully unsupervised approaches promise to train the model directly on the target images without requiring curated pretraining datasets. However, they struggle with the inherent sparsity of meaningful programs in the search space. In this paper, we present the first unsupervised algorithm capable of parsing constructive solid geometry (CSG) images into context-free grammar (CFG) without pretraining via non-differentiable renderer. To tackle the \emph{non-Markovian} sparse reward problem, we combine three key ingredients---(i) a grammar-encoded tree LSTM ensuring program validity (ii) entropy regularization and (iii) sampling without replacement from the CFG syntax tree. Empirically, our algorithm recovers meaningful programs in large search spaces (up to $3.8 \times 10^{28}$). Further, even though our approach is fully unsupervised, it generalizes better than supervised methods on the synthetic 2D CSG dataset. On the 2D computer aided design (CAD) dataset, our approach significantly outperforms the supervised pretrained model and is competitive to the refined model.
\end{abstract}

\section{Introduction}

 Image generation is extensively studied in machine learning and computer vision literature. Vast numbers of papers have explored image generation through
 low-dimensional latent representations \citep{goodfellow2014generative, arjovsky2017wasserstein, li2017mmd, kingma2013auto, van2017neural, oord2016pixel}. However, it is challenging to learn disentangled representations that allow control over each component of the generative models separately ~\citep{higgins2017beta, kim2018disentangling,locatello2018challenging, chen2016infogan}. In this paper, we tackle the problem of CFG program generation from constructive solid geometry (CSG) \citep{hubbard1990constructive} and computer aided design (CAD) images, which are commonplace in engineering and design applications. 
 Parsing a geometric image into a CFG  program not only enables selective manipulations of the desired components while preserving the rest, but also provides a human-readable alternative to the opaque low-dimensional representations generated by neural networks. Our model for extracting the programs can be viewed as an encoder and the renderer that reconstructs the image as a decoder. We focus on a non-differentiable renderer (most design software are non-differentiable, \eg Blender and Autodesk). They are more common than differentiable ones \citep{li2018differentiable,liu2019beyond, kania2020ucsg}, but they are also more challenging to work with neural networks because their discrete nature cannot be integrated into the network and the gradients w.r.t. the rendered images are \emph{inaccessible}.

A common scheme for parsing images (\eg CSG images) into programs (\eg CFG programs) for non-differentiable renderers involves two steps: first use synthetic images with ground-truth programs for supervised pretraining, followed by REINFORCE fine-tuning \citep{sharma2018csgnet, ellis2019write} on the target image dataset. Sampling programs from a grammar can provide data suitable for supervision if the target images are restricted to combinations of geometric primitives specified in the grammar.  There are two limitations of supervised methods. Firstly, it maximizes the likelihood of a single reference program while penalizing many other correct programs \citep{bunel2018leveraging} using maximum likelihood estimation (MLE). This observation is known as \emph{program aliasing} and it adversely affects supervise learning's performance. Secondly, it does not generalize well to the test images not generated by the grammar. 
REINFORCE fine-tuning is proposed to remedy the two limitations \citep{bunel2018leveraging, sharma2018csgnet}. The transition between the supervised pretraining and REINFORCE fine-tuning is delicate, however, because the model is sensitive to bad gradient updates that cause the grammar structure to collapse among the generated programs. Additionally, the quality of the curated pretraining dataset can limit the downstream model's ability to generalize.

In this paper, however, we focus on the more interesting and more challenging \emph{unsupervised task}, when ground-truth programs of the images are not available for training.
Compared with the supervised pretrained alternative, 
unsupervised image parsing is under explored. The benefits of an unsupervised approach are:
\begin{itemize}
    \item \textbf{Train directly on the target domain.} Instead of training on a curated synthetic dataset that mimic the target dataset, \eg combinations of geometric shapes specified in the CFG imitating a CAD dataset exceeding the primitives of the grammar, we can train directly on the target images. The synthetic dataset may not reflect the target image dataset accurately and can subsequently lead to insufficient generalization during test time.
    \item \textbf{Treat all correct programs equally.} When multiple programs correspond to the reconstruction of the same image, while supervised methods only optimize for programs included in the synthetic ground-truth dataset, the equivalent programs receive equal rewards under RL. 
\end{itemize}

Despite the benefits, designing an approach without supervised pretraining is challenging. Because of the problem's discrete nature, we rely on tools from reinforcement learning, such as REINFORCE. However, the program space grows exponentially with the length of the program and valid programs are too sparse in the search space to be sampled frequently enough to learn.
Training with the naive REINFORCE provides no performance gain in our experiments.  RL techniques such as Hindsight Experience Replay \citep{andrychowiczhindsight} that mitigate the sparse reward problem cannot be applied here due to the Markov assumption on the model. We improve the sample efficiency of the REINFORCE algorithm and show that our improved approach achieves competitive results to a two-step model. We further demonstrate that our method generalizes better on a synthetic 2D CSG dataset than a supervised method. On a 2D CAD dataset, which was \emph{NOT} generated by CFG and thus cannot be captured sufficiently by a synthetic dataset, our method exceeds the results of a pretrained model by a large margin and performs competitively to the refined models.

Here we summarize the key ingredients that help successfully learn to parse an image \emph{without program supervision} while using a non-differentiable renderer \emph{without direct gradient propagation} w.r.t. the rendered images:
\begin{itemize}
    
    \item We incorporate a \textbf{grammar-encoded tree LSTM} to impose a structure on the search space such that the algorithm is sampling a path in the CFG syntax tree top-down. This guarantees  validity of the output program.
    
    \item We propose an \textbf{entropy estimator} suitable for sampling top-down from a syntax tree to encourage exploration of the search space by entropy regularization. 
    
    \item Instead of relying on naive Monte Carlo sampling, we adopt \textbf{sampling without replacement} from the syntax tree to obtain better entropy estimates and REINFORCE objective for faster convergence.
    
\end{itemize}

 \section{Related Work}

 
Program synthesis attracts growing interests from researchers in machine learning. Supervised training is a natural choice for input/output program synthesis problems \citep{parisotto2016neuro, chen2018execution, devlin2017robustfill, yin2017syntactic, balog2016deepcoder, zohar2018automatic}. \citet{shin2018improving} use the input/output pairs to learn the execution traces. \citet{bunel2018leveraging} use RL to address program aliasing, however, supervised pretraining is still necessary to reap its benefits. Approaches to ensure valid outputs involve syntax checkers \citep{bunel2018leveraging} or constructions of abstract syntactic trees (AST) \citep{parisotto2016neuro, yin2017syntactic, kusner2017grammar, chen2018tree}. A graph can also model the information flow in a program \citep{brockschmidt2018generative}. 

Vision-as-inverse-graphics focuses on parsing a scene into a collection of shapes or 3D primitives, \eg cars or trees, with parameters, \eg colors or locations, that imitates the original scene \citep{tulsiani2017learning, romaszko2017vision, wu2017neural}. \citet{yao20183d} further manipulate the objects de-rendered, such as color changes. Stroke-based rendering creates an image similar to how we write and draw. Some of the examples are recreating paintings by imitating a painter's brush strokes \citep{huang2019learning}, drawing sketches of objects \citep{sketchrnn}. SPIRAL \citep{ganin2018synthesizing} is an adversarially trained deep RL agent that can recreate MNIST digits and Omniglot characters. Contrary to our problem, a grammar structure is unnecessary to both vision-as-inverse-graphics and stroke-based rendering. 

Research on converting images to programs relates more closely to our work \citep{sharma2018csgnet, ellis2019write, ellis2018learning, liu2018learning, shin2019synthetic, beltramelli2018pix2code, kania2020ucsg}. \citet{tian2019learning, kania2020ucsg} incorporate differentiable renderers into the learning pipeline while we treat our renderer as an external process independent from the learning process, thus unable to propagate gradient through the renderer. Furthermore, \cite{kania2020ucsg} construct the parse tree bottom up and pre-determines the number of leaves, as opposed to top down like ours which is more general. \citet{ellis2018learning} use neural networks to extract shapes from hand-drawn sketches, formulate the grammatical rules as constraints and obtain the final programs by optimizing a constraint satisfaction problem. Similarly, \citet{du2018inversecsg} cast the problem of parsing a 3D model into a CSG tree as a constraint satisfaction problem and solve by an existing SAT solver. This process can be computationally expensive compared to neural network based solutions. More relevantly, \cite{sharma2018csgnet} perform program synthesis by supervised pretraining before RL fine-tuning to generalize to a CAD dataset. \citet{ellis2019write} pretrain a policy with supervision from synthetic data and learns a value function with REINFORCE. Both are used for pruning unpromising candidates during test time. Their reward function is binary and cannot approximate images not generated by a grammar, as opposed to our model.  

 
\section{Proposed Algorithm} \label{proposed algo}
\paragraph{CSG Image and CFG Program.} We use constructive solid geometry (CSG) ~\citep{hubbard1990constructive} to describe an image. The input of our model are images constructed from geometric shapes (\eg square, circle, ...) each with a designated size and location (see Figure~\ref{fig:shape_encoding}). The outputs of the model are context-free grammar (CFG) programs. 
In the CFG specification \citep{sharma2018csgnet}, $S$, $T$, and $P$ are non-terminals for the start, operations, and shapes. The rest are terminals, \eg $+$ (union), $*$ (intersection), $-$ (subtraction), and $c(48,16,8)$ stands for a circle with radius $8$ at location $(48,16)$ in the image. Figure~\ref{fig: qualitative & example} contains examples of CSG images and their corresponding programs. Each line below is a \emph{production rule} or just \emph{rule} for simplicity:

\begin{align} \label{rule1}
    S &\rightarrow E \\ \label{rule2}
    E &\rightarrow EET | P \\ \label{rule3}
    T &\rightarrow + | - | * \\ \label{rule4}
    P &\rightarrow SHAPE_1 | SHAPE_2 | \cdots | SHAPE_n.
\end{align}


\subsection{Learning with REINFORCE}

\begin{figure*}
    \centering
    \includegraphics[width=\linewidth]{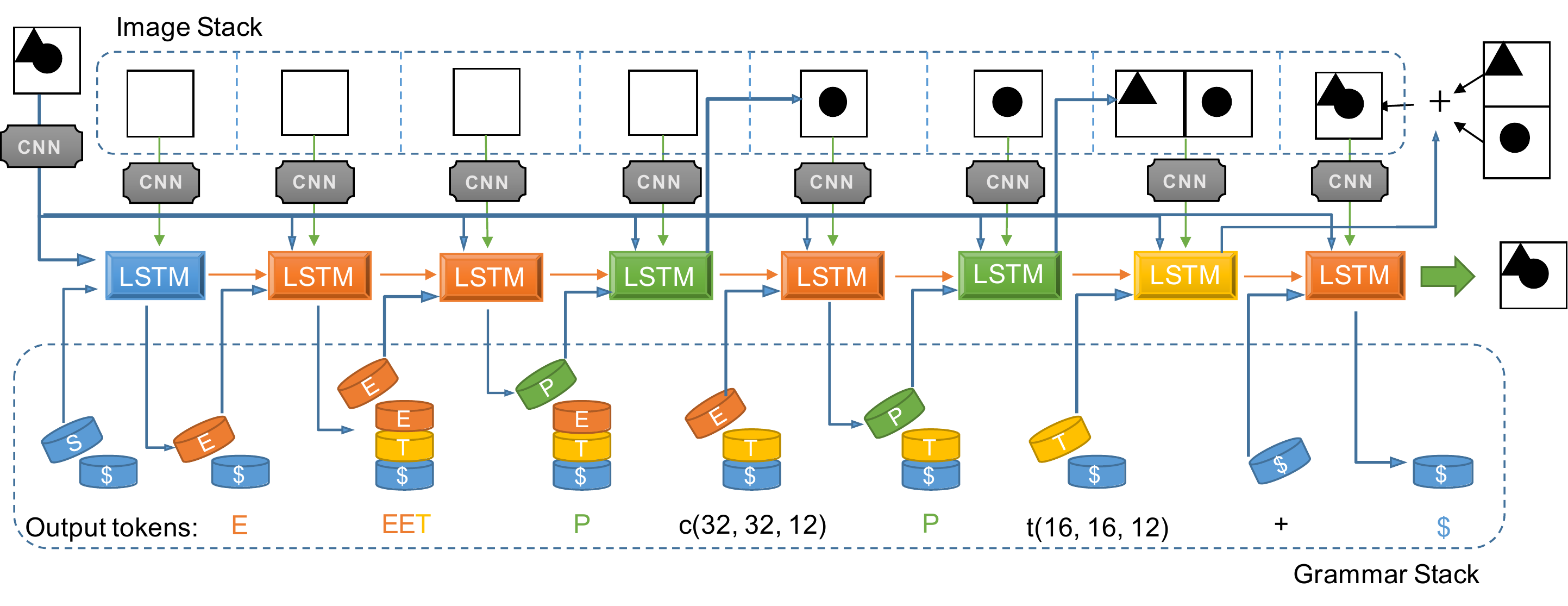}
    \vspace{-10pt}
    \caption{This is an example of the grammar encoded tree LSTM at work. The top layer of images demonstrates the image stack and the bottom layer demonstrates the grammar stack. The blue, orange, yellow and green colored LSTM cell generates grammatical tokens according to the CFG rule \ref{rule1}, \ref{rule2}, \ref{rule3} and \ref{rule4} respectively. In implementation, we can constrain the output space by adding a mask to the output of the LSTM and render the invalid options with close to zero probability of being sampled. }
    \label{fig:tree_lstm}
\end{figure*}

A model trained with the REINFORCE objective only \citep{sharma2018csgnet} is unable to improve beyond the lowest reward (Section \ref{abalation}). We propose three components to enable the RL approach to learn in this setting: (i) a tree model entropy estimator to encourage exploration; (ii) sampling without replacement in the program space to facilitate optimization and further encourage exploration; (iii) a grammar-encoded tree LSTM to ensure valid output sequences with an image stack to provide intermediate feedback. We start this section by discussing objective and reward function.

\textbf{Objective Function} Our model consists of a CNN encoder for input images, an embedding layer for the actions, and an RNN for generating the program sequences (see Figure \ref{fig:tree_lstm}). The model is trained with entropy regularized REINFORCE \citep{williams1992simple}. Here, let $\mathcal{H}(s)$ and $f(s)$ stand for the entropy (we will define this later) and reward function of sequence $s$, respectively, and let $\theta$ denote the parameters of the model. The objective is optimized as follows

\begin{equation} \label{eq: entropy}
    \Delta \theta \propto  \mathbb{E}_{s \sim P_\theta (s)} [ \nabla_\theta \log P(s) f(s) ] + \alpha \nabla_\theta \mathcal{H}(s)
\end{equation}

\textbf{Reward Function} The output program $s$ is converted to an image $\mathbf{y}$ by a non-differentiable renderer. The image is compared to the target image $\mathbf{x}$ and receives a reward $f(s)= R(\mathbf{x}, \mathbf{y})$. We use Chamfer Distance (CD) as part of the reward function. The CD calculates the average matching distance to the nearest feature and is a greedy estimation of image similarity, unlike Optimal Transport (OT). However, OT is not computationally feasible for RL purpose. 

\begin{algorithm}[H]
\caption{Sampling w/o Replacement Tree LSTM }
\label{alg:beam search tree lstm}
\textbf{Input:} \; \text{Target Image $\mathbf{x}$}, Number of samples $k$ \\
\textbf{Initialize:} \; \text{Grammar stack $S$, Image stack $I$, Sample set $\mathbb{B}$} 
\begin{algorithmic} 
\State $\text{Encode the target image} \; \tilde{T} \leftarrow \mathrm{Encode}(T)$
\State $\mathbb{B} = \{\mathbf{s}^i_0, G_{s^i_0}, \phi_{i,0}| \mathbf{s}^i_0 = \emptyset, G_{s^i_0} = 0, \phi_{i,0} = 0\} \; \text{for} \; i \in {1, 2, \cdots k + 1} \; \text{and} \; \mathcal{H}(v) = 0$

\For {$j := 1 \; \text{ to} \; n$}

    \State $\bm{\Phi}_{:, j}, \mathcal{H}_{i, j} \leftarrow \mathrm{TreeLSTM}(S, I, \mathbb{B}, \mathbf{x})$ (Algorithm \ref{alg:tree lstm})
    \State $\mathbb{B} \leftarrow \mathrm{Sample\_w/o\_Replacement}(\bm{\Phi}_{:, {j}},\mathbb{B}, k)$ 
    \State \text{(See Algorithm ~\ref{alg:beam search} and ~\citep{kool2019stochastic} )}
    \State $\hat{\mathcal{H}}_D \leftarrow \hat{\mathcal{H}}_D + \sum^n_{j=1} \frac{1}{W_j(S)} \sum_{s^i \in S} \frac{p_\theta (s^i_j)}{q_{\theta, \kappa} (s^i_j)} \mathcal{H}_{i, j}  \; $
    \State \text{(Equation \ref{eq: BS beam search entropy} in Appendix~\ref{sampling without replacement})}
    \State \textbf{if} $s^i_{j} \in \mathcal{G} $ \textbf{then} $S_i.\mathrm{push}(s^i_{j})$ 
    \State\textbf{else} $I_i.\mathrm{push}(s^i_{j}), \forall s^i \in \mathbb{B}$
    \EndFor
    
\State $\mathbf{y}_i = \mathrm{Render}(s^i) \; \text{for} \; i \in {1, 2, \cdots k}$ 
\State $\text{Maximize} \;  \mathbb{E} [ \sum^k_{i=1} R(\mathbf{x}, \mathbf{y}_i) ] + \alpha \hat{\mathcal{H}}_D$ 
\end{algorithmic}
\end{algorithm}

Formally, let $x \in \mathbf{x}$ and $y \in \mathbf{y}$ be pixels in each image respectively. Then the distance $Ch(\mathbf{x}, \mathbf{y})$ is
\begin{equation} \label{eq: schamfer}
    Ch(\mathbf{x}, \mathbf{y}) = \frac{1}{ |\mathbf{x}|} \sum_{x \in \mathbf{x}} \min_{y \in \mathbf{y}} ||x - y||_2 + \frac{1}{\|\mathbf{y}\|} \sum_{y \in \mathbf{y}} \min_{x \in \mathbf{x}} ||x - y||_2
\end{equation}

The CD is scaled by the length of the image diagonal ($\rho$) \citep{sharma2018csgnet} such that the final value is between 0 and 1. For this problem, the reward $1 - Ch(\mathbf{x}, \mathbf{y})$ mostly falls between 0.9 and 1. We exponentiate $1 - Ch(\mathbf{x}, \mathbf{y})$ to the power of $\gamma = 20$ to achieve smoother gradients \citep{laud2004theory}. We add another pixel intersection based component to differentiate shapes with similar sizes and locations. The final reward function is defined as: 
\begin{equation}\label{eq: rewardfunction}
    R(\mathbf{x}, \mathbf{y}) = 
    \max(\delta, (1 - \frac{Ch(\mathbf{x}, \mathbf{y})}{\rho})^{\gamma} + \frac{\sum_{x \in \mathbf{x} \cap \mathbf{y}} 1}{\sum_{ x \in \mathbf{x}}1}) \end{equation}
 The first and second part of the reward function provide feedback on the physical distance and similarity of the prediction respectively.  We clip the reward below $\delta = 0.3$ to simplify it when the quality of the generated images are poor. A low reward value provides little insight on its performance and is largely dependent on its target image. Similar reward clipping idea was used in DQN \citep{mnih2013playing}. 

\subsection{Exploration with Entropy Regularization} \label{exploration}


Entropy regularization in RL is a standard practice for encouraging exploration. Here we propose an entropy estimation for sampling top-down from a syntax tree. Let $S$ denote a random variable of possible programs. Its entropy is defined by $\mathcal{H}(S) = \mathbb{E}[-\log P(S)]$ \footnote{Following Rule \eqref{rule1} and Rule \eqref{rule2} we overload $S$ and $P$.}. Estimating $\mathcal{H}(S)$ is challenging because the possible outcomes of $S$ is exponentially large and we cannot enumerate all of them. Given the distribution $P$, the naive entropy estimator is
\vspace{-5pt}
\begin{equation} \label{eq:naive entropy}
    \hat{\mathcal{H}} = -\frac{1}{K}\sum_{i=1}^K \log P(s^i),
\end{equation}
where $\{s^i\}_{i=1}^K$ are iid samples from $P$. In practice, when $K$ is not exponentially large, this estimator has huge variance. We provide an improved estimator designed for our syntax tree: First, we decompose the program $S$ into $S=X_1 \dots X_n$, where each $X_j$ is the random variable for the token at position $j$ in the program. Under autoregressive models (\eg RNN), we can access the conditional probabilities, and this allows us to construct decomposed entropy estimator $\hat{\mathcal{H}}_D$ as 
\vspace*{-5pt}
\begin{equation} \label{eq: stepwise entropy}
    \hat{\mathcal{H}}_D = \frac{1}{K} \sum^K_{i=1} \sum^n_{j=1} \mathcal{H}(X_j |  X_1 = x_1^i, \cdots, X_{j-1} = x_{j-1}^i), 
\end{equation}
\vspace{-5pt}

where $s^i = x_1^i, \dots, x_n^i$, and $\mathcal{H}(X_j |  X_1 = x_1^i, \cdots, X_{j-1} = x_{j-1}^i)$ is the conditional entropy. The below lemma states that $\hat{\mathcal{H}}_D$ is indeed an improved estimator over $\hat{\mathcal{H}}$.
\begin{lemma} \label{entropy lemma}
The proposed decomposed entropy estimator $\hat{\mathcal{H}}_D$ is unbiased with lower variance, that is $\mathbb{E}[\hat{\mathcal{H}}_D] = \mathcal{H}(S)$ and  $\mbox{Var}(\hat{\mathcal{H}}_D) \leq \mbox{Var}(\hat{\mathcal{H}})$.
\end{lemma}

The proof can be found in the Appendix \ref{unbiasness} and  \ref{lowvariance}.

\begin{figure*}
    \centering
    \includegraphics[width=\linewidth]{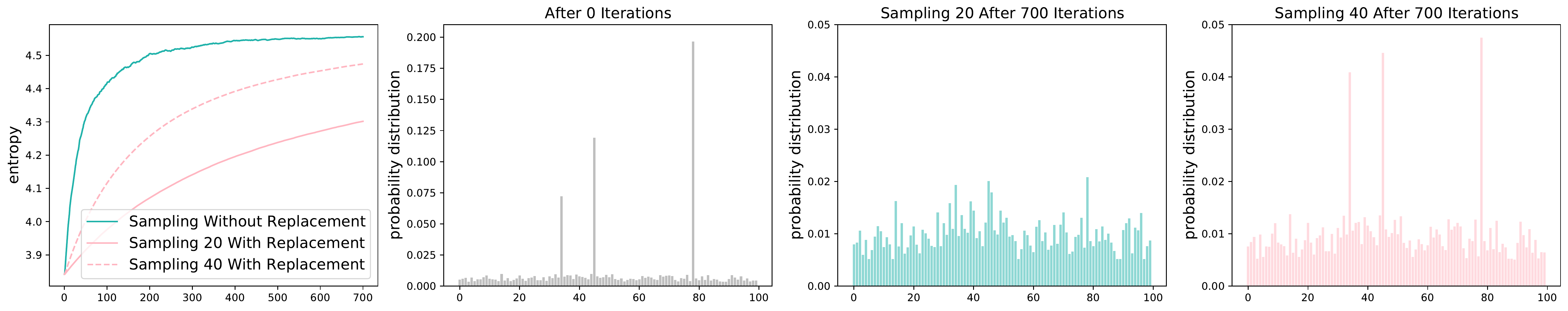}
    \caption{ The left most image demonstrates the entropy value increases over 700 iterations by sampling 20 distinct samples with and without replacement as well as sampling 40 samples with replacement. The second image shows the initial distribution. The third and fourth images show the final distributions.  }
    \vspace{-10pt}
    \label{fig:entropy_replacement}
\end{figure*}

\begin{algorithm}[H] 
\caption{Sampling\_w/o\_Replacement }
\label{alg:beam search}
\begin{algorithmic}
\State {\bfseries Input:} \text{ Log probability at time $j$ $\bm{\Phi}_{:, j}$ Beam Set $\mathbb{B}$, } 
\State \; \; \; \; \; \text{Number of beams $k$}\\
\Function{Sampling\_w/o\_Replacement}{$\bm{\Phi}_{:, j},\mathbb{B},k$}
    \State $\Tilde{\bm{G}}_j \leftarrow \emptyset$
    \For{$\mathbf{s}^i_{j-1}, G_{s^i_{j-1}} \in \mathbb{B} \; \text{and} \; \vec{\bm{\phi}}_{i, j} \in \bm{\Phi}_{:, j}$}
    \State $\vec{\mathbf{G}}_{\bm{\phi}_{i, j}} \sim \text{Gumbel}(\vec{\bm{\phi}}_{{i, j}})$ 
    \State $ Z_{i, j} \leftarrow \text{max}(\vec{\mathbf{G}}_{\bm{\phi}_{{i, j}}})$
    \State Calculate $\vec{\tilde{\mathbf{G}}}_{\bm{\phi}_{{i, j}}}$ (Equation \ref{eq: G prime} in Appendix \ref{sampling without replacement})
    \State \text{Aggregate the values in the vector  $\Tilde{\mathbf{G}}_{\phi_{{i, j}}}$} 
    \State $\Tilde{\bm{G}}_j \leftarrow \Tilde{\bm{G}}_j \cup \Tilde{\mathbf{G}}_{\phi_{{i, j}}}$
    \EndFor
    \State $\text{Choose top $k + 1$ values in $\Tilde{\bm{G}}_j \in \mathbb{R}^{(k+1)\cdot A}$ }$
    \State $\text{and form the new beam set }$ 
    \State $ \tilde{\mathbb{B}} = \{\mathbf{s}^{i}_j, G_{s^i_j}, \phi^i_j | \mathbf{s}^{i}_{j} = \tilde{\mathbf{s}}^{i}_{j-1} \cup \tilde{s}^{i}_{j}, G_{s^i_j} = \tilde{G}_{i, j}, \phi_{i,j} = \log P(\mathbf{s}^i_j)  \} $ \text{where} $\; i \in 1, 2, \cdots, k + 1$
    \State \textbf{return} $\tilde{\mathbb{B}}$
\EndFunction
\end{algorithmic}
\end{algorithm}

\subsection{Effective Optimization By Sampling Without Replacement} \label{swr}


After establishing our REINFORCE method with entropy regularization objective, now we show the intuition behind choosing sampling without replacement (SWOR) over sampling with replacement (SWR). For this explanation, we use a synthetic example  (Figure \ref{fig:entropy_replacement}).

We initialize a distribution of $m = 100$ variables with three of them having significantly higher probability than the others (Figure \ref{fig:entropy_replacement} (2)). The loss function is entropy. Its estimator is $\frac{1}{m}\sum^m_{i=1} \log p_i$ for SWR and $\sum^m_{i=1} \frac{p_i}{q_i} \log p_i$ for SWOR. In both cases, $p_i$ is the $i$-th variable's probability. $q_i$ is the re-normalized probability after SWOR. $\frac{p_i}{q_i}$ is the importance weighting. The increase in entropy by sampling 20 variables without replacement is more rapid than 40 variables with replacement. At the end of the 700 iterations, the distribution under SWOR is visibly more uniform than the other. SWOR would achieve better exploration than SWR.

To apply SWOR to our objective, the REINFORCE objective and the entropy estimator require importance weightings. Let $s^i_j$ denotes the first $j$ elements of sequence $s^i$:
\begin{equation}\label{eq: vanilla reinforce}
\nabla_\theta \mathbb{E}_{s \sim p_\theta (s)} [ f(s) ] \approx \sum_{s^i \in S} \frac{\nabla_\theta p_\theta (s^i)}{q_{\theta} (s^i)} f(s^i) \; \; \; \text{and, }
\end{equation} 
\vspace{-10pt}
\begin{align} \label{eq: vanilla beam search entropy}
    &\hat{\mathcal{H}}_D \approx  \sum^n_{j=1}  \sum_{s^i \in S} \frac{p_\theta (s^i_j)}{q_{\theta} (s^i_j)} \mathcal{H}(X_j | X_1 = x^i_1, \cdots, 
     X_{j-1} = x^i_{j-1})
\end{align}

Implementing SWOR on a tree structure to obtain the appropriate set of programs $S$ is challenging. It is not practical to instantiate all paths and perform SWOR bottom-up. Instead, we adopt a form of stochastic beam search by combining top-down SWOR with Gumbel trick that is equivalent to SWOR bottom-up \citep{kool2019stochastic}.  The sampling process is described in Algorithm \ref{alg:beam search}. For more detailed explanation, the implementation of the re-normalized probability $q_\theta(s^i)$ as well as some additional tricks of variance reduction for the objective function, please refer to Appendix \ref{sampling without replacement}.

\subsection{Grammar Encoded Tree LSTM}



\begin{figure}%
    \centering
    \subfloat[Example output of each method]{{\includegraphics[width=\linewidth]{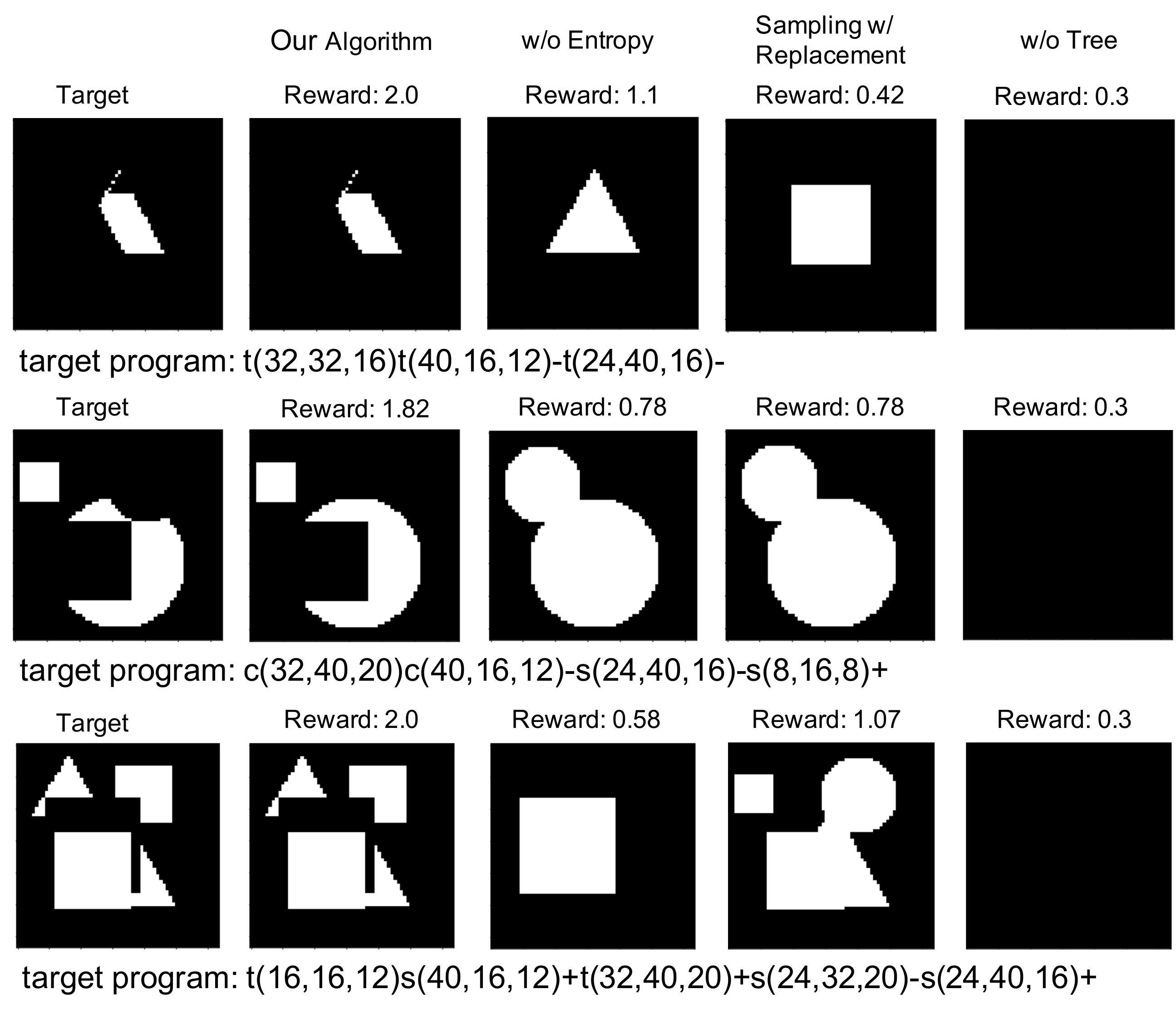} }}%
    \qquad
    \subfloat[Example outputs of sampling without replacement]{{\includegraphics[width=\linewidth]{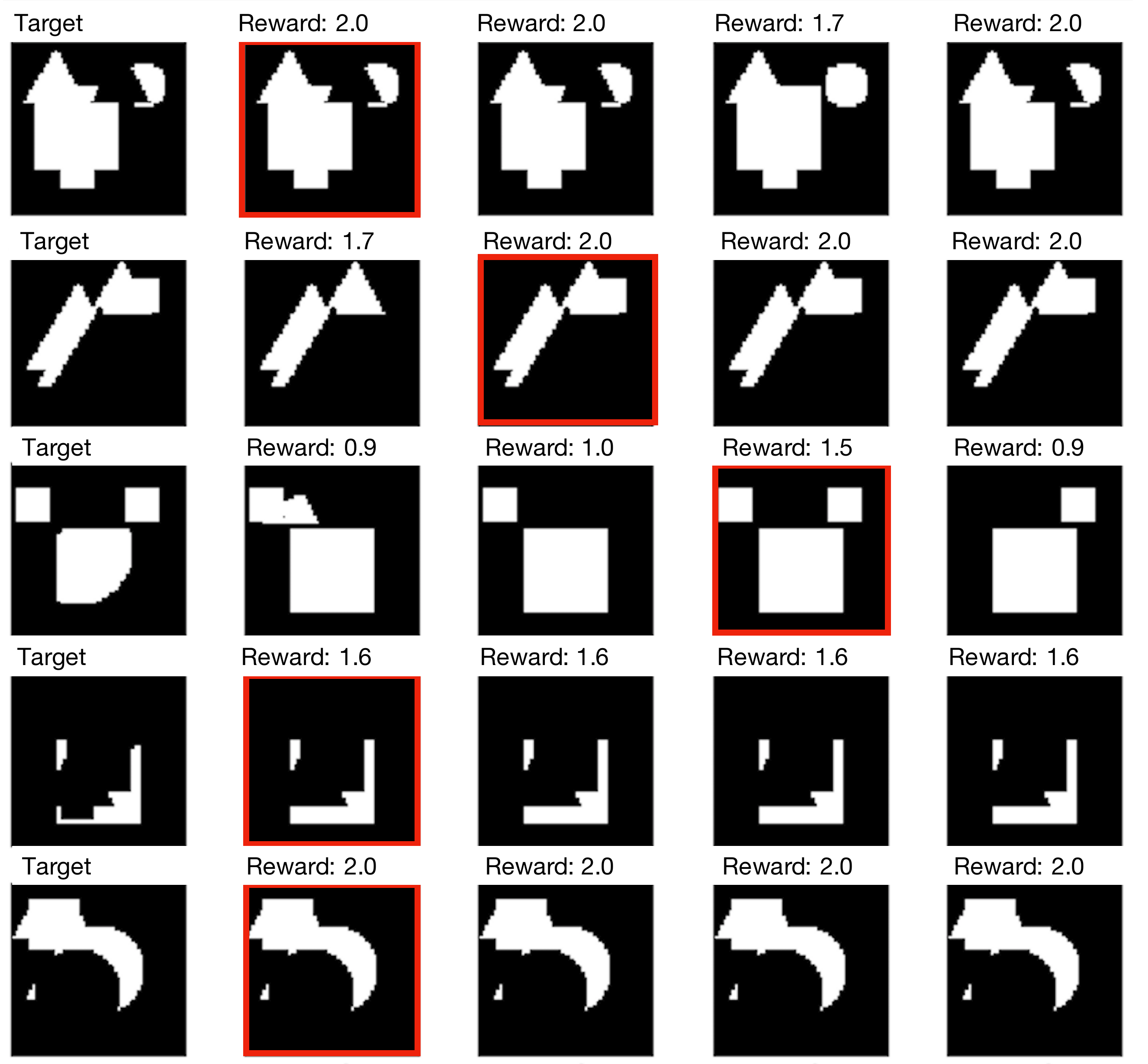} }}%
    \caption{(a) We show a target image from each dataset and attach its correct program below. To the right are the reconstructed output programs from our algorithm and three variants each removing one design component. The reward is on top of the reconstructed images. (b) Some reconstructed example output programs of our algorithm. Each row represents one data point. The leftmost images of the five columns are the target images and the four columns to their right are the reconstructed outputs of four samples. The final output highlighted in red has the highest reward.}%
    \vspace*{-20pt}
    \label{fig: qualitative & example}
\end{figure}

We introduce a \emph{grammar-encoded tree LSTM} which encodes the production rules into the model, thus guaranteed to generate correct programs, and significantly reduce the search space during the training \citep{kusner2017grammar, alvarez2016tree,  parisotto2016neuro, yin2017syntactic}. There are 3 types of production rules in the grammatical program generation -- shape selection ($P$), operation selection ($T$), and grammar selection ($E$). Grammar selection in this problem setting includes $E\rightarrow EET$, and $E\rightarrow P$ and they decide whether the program would expand. We denote the set of shape, operations and non-terminal outcomes (\eg $EET$ in Rule \eqref{rule2}) to be $\mathcal{P}$, $\mathcal{T}$ and $\mathcal{G}$ respectively.
A naive parameterization is to let the candidate set of the LSTM output to be $\{S,\$\} \cup \mathcal{T} \cup \mathcal{P}$, where $\$$ is the end token, and treat it as a standard language model to generate the program~\citep{sharma2018csgnet}. The model does not explicitly encode grammar structures, and expect the model to capture it implicitly during the learning process.  The drawback is that the occurrence of valid programs is sparse during sampling and it can prolong the training process significantly. 
\begin{algorithm}[H] 
\caption{TreeLSTM Model }
\label{alg:tree lstm}
\begin{algorithmic}
\State {\bfseries Input:} \text{Grammar Stack $S$, Image Stack $I$,} 
\State \; \; \; \; \text{Target Image $\mathbf{x}$, Sample Set $\mathbb{B}$} \\
\Function{TreeLSTM}{$S, I, \mathbb{B}, \mathbf{x}$}
    \State $ \tilde{\mathbf{x}} \leftarrow \mathrm{Encode}(\mathbf{x})$
    \For { $s^i_{j-1}, \phi_{i,{j-1}} \in \mathbb{B}$} 
        \State $ g_i \leftarrow S_i.\mathrm{pop}()$ 
        \State  $ \tilde{g}_i \leftarrow \mathrm{Embed}(g_i)$
        \State  $ \tilde{I}_i \leftarrow \mathrm{Encode}(I_i)$
        \State  $H_{i,j} \leftarrow \mathrm{LSTM}(\tilde{g}_i,\tilde{I}_i,\tilde{\mathbf{x}}, H_{i,j-1})$
        \State  $\mathbf{p_{i,j}} \leftarrow \mathrm{softmax}(f(H_{i,j}) + \mathrm{Mask}(g_i))$
        \State  $\text{Estimate entropy at this level:} $
        \State  $\mathcal{H}_{i, j} = \mathbf{\vec{p}_{i,j}} \boldsymbol{\cdot} \log \mathbf{\vec{p}_{i,j}}$
        \State $\text{Update the log probabilities of partial sequences } $
        \State $\vec{\bm{\phi}}_{{i, j}} =  \vec{\mathbbm{1}} \cdot \phi_{i,{j-1}}  + \log \mathbf{p_{i, j}} $
        
    \EndFor
    \State \textbf{\normalsize return} $\bm{\Phi}_{:, j}, \mathcal{H}_{i,j} $
\EndFunction
\end{algorithmic}
\end{algorithm}



The proposed model can be described as an RNN model with a \emph{masking mechanism} by maintaining a grammar stack to rule out invalid outputs. We increase the size of the total output space from $2+|\mathcal{P}| + |\mathcal{T}|$ of the previous approach (\eg \citep{sharma2018csgnet}) to $2+|\mathcal{P}| + |\mathcal{T}| + |\mathcal{G}|$ by including the non-terminals. During the generation, we maintain a stack to trace the current production rule. Based on the current non-terminal and its corresponding expansion rules, we use the masking mechanism to weed out the invalid output candidates. Take the non-terminal $T$ for example, we mask the invalid outputs to reduce the candidate size from $2+|\mathcal{P}| + |\mathcal{T}| + |\mathcal{G}|$ to $|\mathcal{T}|$ only. In this process, the model will produce a sequence of tokens, including grammatical, shape and operation tokens. We only keep the terminals as the final output program and discard the rest. The resulting programs are ensured to be grammatically correct. During the generation process, grammatical tokens are pushed onto the grammar stack while intermediate images and operations are pushed onto an \emph{image stack}. Images in the image stack are part of the input to the LSTM to aid the inference in the search space. A step-by-step guide through the tree LSTM for better understanding is in Appendix \ref{grammar tree lstm tracing} and Figure \ref{fig:tree_lstm} is a visual representation of the process.


    

\section{Experiments}

There are two datasets we used for experiments -- a synthetic dataset generated by the CFG specified in Section \ref{proposed algo} and a 2D CAD furniture dataset. We compared the result of our algorithm to that of the same neural network trained with supervision of ground-truth programs. We observed that the supervised model shows poorer generalization in both datasets despite its access to additional ground-truth programs. We provide qualitative and quantitative ablation study of our algorithm on the synthetic dataset. We also showed that our algorithm can \emph{approximate} the CAD images with programs despite not having exact matches. On the CAD furniture dataset, it outperforms the supervised pretrained model and achieves competitive result to the refined model. Additionally, we verify empirically that the stepwise entropy estimator (Equation ~\ref{eq: stepwise entropy}) indeed has smaller variance than the naive estimator (Equation ~\ref{eq:naive entropy}) as proven in Lemma \ref{entropy lemma}. 


\subsection{Synthetic Dataset Study}
\label{synthetic}
We use three synthetic datasets to test our algorithm. The action space includes 27 shapes (Figure \ref{fig:shape_encoding}), 3 operations and 2 grammar non-terminals to create a 64 by 64 images. The search space for an image up to 3 shapes (or program length 5) is around $1.8 \times 10^5$ and it gets up to $1.1 \times 10^9$ for 5 shapes (or program length 9). We separate our dataset by the length of the program to differentiate images with increasing complexity. Our synthetic dataset is generated by filtering out the duplicates and empty images in combinations of shape and operation actions in text. Images are considered duplicates if only 120 pixels are different between the two and are considered empty if there are no more than 120 pixels in the image. Table~\ref{table: dataset size} contains the dataset size information. 
\subsubsection{ Ablation Study of Design Components}
\label{abalation}
\begin{figure*}
    \centering
    \includegraphics[width=\linewidth]{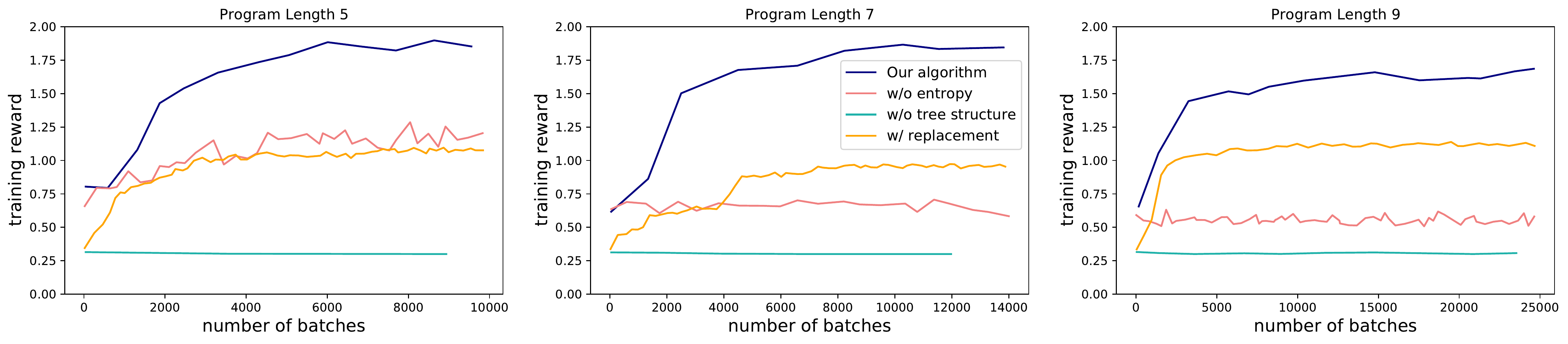}
    \caption{From left to right, we have reward per batch for programs of length 5, 7 and 9. It demonstrates the performance of our algorithm and controlled comparison in performance with alternative algorithms by removing one component at a time.
   }
    \label{fig:performance}
\end{figure*}

For these 3 datasets, we sampled 19 programs without replacement for each target image during training. The negative entropy coefficient is 0.05 and the learning rate is 0.01. We use SGD with 0.9 momentum.


Removing either one of the three design components has reduced the performance of our algorithm. Under sampling with replacement setting, the model is quickly stuck at a local optimum (Figure \ref{fig:performance} (yellow)). 
Without the entropy term in the objective function, the reward function is only able to improve on the length 5 dataset but fails to do so on longer programs. Both techniques have facilitated exploration that helps the model to escape local minimum. 
Without tree structure, the reward stays around the lowest reward (Figure \ref{fig:performance} (green)) because the program is unable to generate valid programs. Grammar encoded tree LSTM effectively constraints the search space such that the sampled programs are valid and can provide meaningful feedback to the model.

We allow variations in the generated programs as long as the target images can be recovered, thus we evaluate the program quality in terms of the reconstructed image's similarity to the target image. We measure our converged algorithm's performance (Table \ref{table: test result}) on the three test sets with Chamfer and IoU reward metrics (Equation \ref{eq: rewardfunction} first and second term). The perfect match receives 1 in both metrics. Figure \ref{fig: qualitative & example} provides some qualitative examples on the algorithms. 

\begin{table}[]
    \centering
    \resizebox{\columnwidth}{!}{%
    \begin{tabular}{||c c c c c||}
     
     \hline
      Type & Length 5 & Length 7 & Length 9 & 2D CAD  \\    
     \hline\hline
     Training set size  & 3600 & 4800 & 12000 & 10000 \\ 
     \hline
     Testing set size & 586 & 789 & 4630 & 3000 \\
     \hline
   \end{tabular}
   }
    \caption{Dataset statistics. }
    \label{table: dataset size}
\end{table}


\subsubsection{Comparison With Supervised Training} \label{supervised learning}

We compare a supervised learning method's train and test results on the synthetic dataset, using the same neural network model as in the unsupervised method. The input at each step is the concatenation of embedded ground truth program and the encoded final and intermediate images. We use the same Chamfer reward metric as in Table \ref{table: test result} to measure the quality of the programs. The test results of the supervised method worsen with the increasing complexity (program length) while the train results are almost perfect across all three datasets. The unsupervised method receives consistently high scores and generalizes better to new data in comparison to the supervised method (Table \ref{table: supervised result}). This phenomenon can be explained by program aliasing \citep{bunel2018leveraging}. The RL method \emph{treats all correct programs equally} and directly optimizes over the reward function in the image space while the supervised method is limited to the content of the synthetic dataset and only optimizes over the loss function in the program space. 

\subsubsection{Supervised Pretrained On Limited Data With REINFORCE Fine-tuning} \label{limited data drop}

\begin{figure}[tb]
    \centering
    \includegraphics[width=\linewidth]{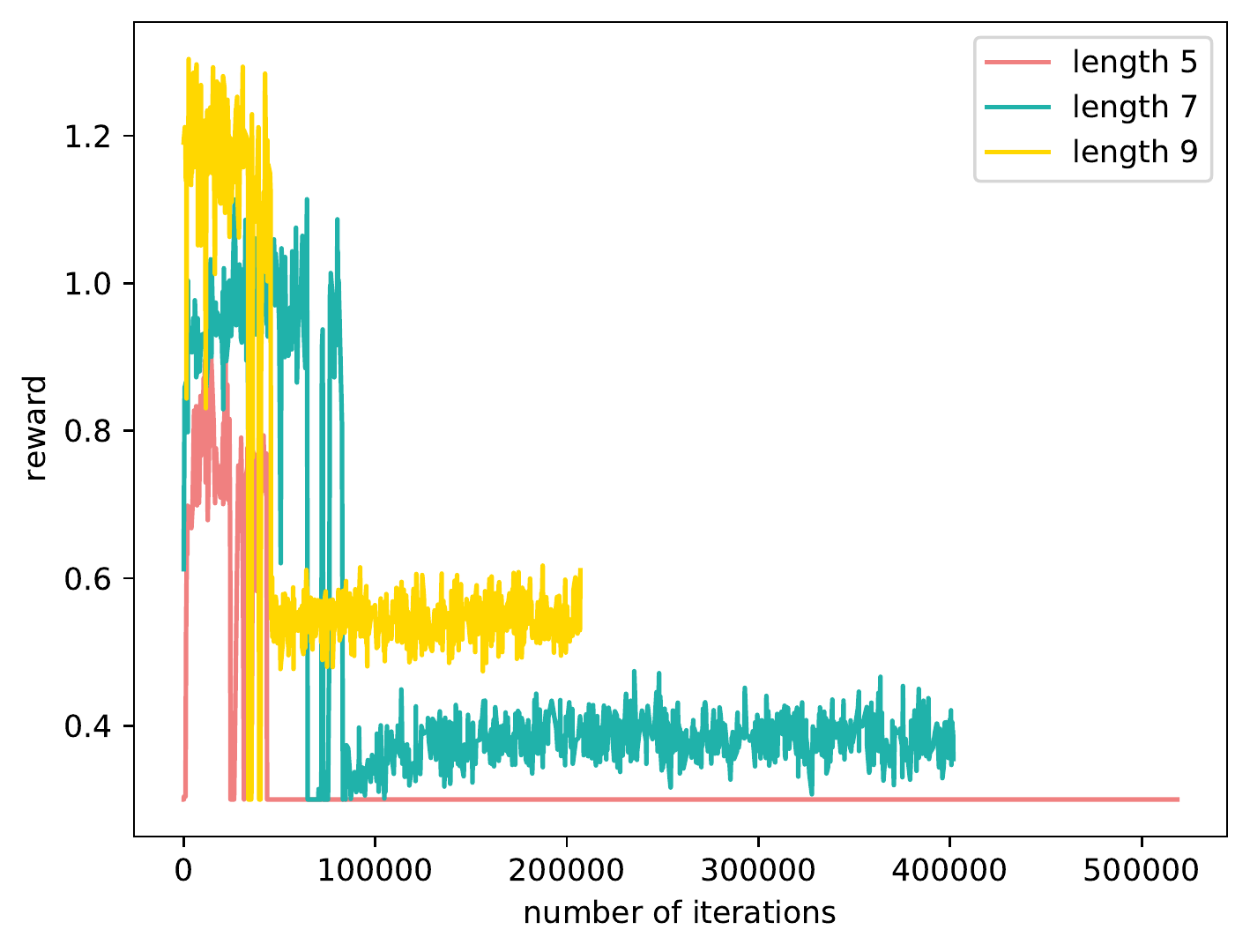}

    \caption{ REINFORCE fine-tuning with pretrained model. }
    \label{fig:supervised_pretrain_RL}
\end{figure}

In this experiment we pretrained the supervised model on a third of the synthetic training dataset till convergence. We take the model and further fine-tune it with vanilla REINFORCE on the full training sets. We report the reward throughout fine-tuning process (Figure \ref{fig:supervised_pretrain_RL}) and it dropped sharply in all three datasets. Our explanation is that while the output of the original supervised pretrained model follow grammatical structures, they are not able to retain the structure consistently after updates during the refinement process, which leads to the collapse.

\subsection{2D CAD Furniture Dataset Study}
\label{2d cad furnature}

\begin{figure*}[!tb]
    
    \centering
    \includegraphics[width=\linewidth]{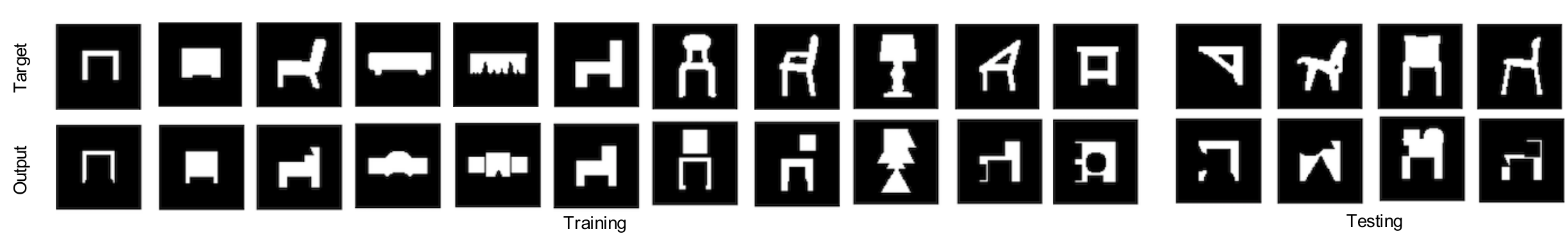}
    \vspace{-20pt}
    \caption{The four examples on the right are from the test set, and the rest on the left are from the training set. The target images are on top and the reconstruction from the output programs are at the bottom.}
    \vspace{-10pt}
    \label{fig: chair}
\end{figure*}

\begin{table}[]
    \centering
\begin{tabular}{||c c c c||}
     \hline
      Test Metric & Length 5 & Length 7 & Length 9  \\ 
     \hline\hline
     Chamfer Reward $\uparrow$ & 0.98 & 0.96 & 0.96  \\ 
     \hline
     IoU Reward $\uparrow$ & 0.99 & 0.96 & 0.96 \\
     \hline
  \end{tabular}
  
  \caption{The performance of the converged model of our algorithm on the test set measured with Chamfer reward and IoU reward.}
  \label{table: test result}
  \vspace{-10pt}
\end{table}
The dataset used in this experiment is a 2D CAD dataset \citep{sharma2018csgnet} that contains binary $64 \times 64$ images of various furniture items. We apply our algorithm to this problem with an action space of 396 basic shapes plus the operations and grammatical terminals described in Section \ref{proposed algo}. We limit the number of LSTM iterations to 24 steps, which corresponds to a maximum of 6 shapes. For an image up to 6 shapes the search space is  $9.4 \times 10^{17}$. If we remove the grammar-encoded tree structure, the search space is $3.8 \times 10^{28}$. In order to scale up to such a big search space from the synthetic experiment, we increase the number of programs sampled without replacement to 550. The higher number usually corresponds to faster convergence and better performance at convergence but the performance gain diminishes at around 500 for this problem. The learning rate and entropy values used are 0.01 and 0.007 respectively. We train the model with only the Chamfer reward (first part of the Equation \ref{eq: rewardfunction}) because these images are not generated by CFG and exact matching solutions do not exist. During training, the reward converges to 0.72. Qualitative results are reported 
from the train and test set (Figure \ref{fig: chair}). The program reconstructions are able to capture the overall profile of the target images. However, the cutouts and angles deviate from the original because 
the shape actions consist solely of unrotated squares, perfect circles and equilateral triangles.

\begin{table}[]
    \centering
\begin{tabular}{||c c c c||}
     \hline
      Chamfer Reward $\uparrow$ & length 5 & length 7 & length 9  \\ 
     \hline\hline
     Training & 1.00 & 1.00 & 0.99\\
     \hline
     Testing & 0.99 & 0.91 & 0.83 \\
     \hline
   \end{tabular}
  \caption{Supervised training results.}
  \label{table: supervised result}
\end{table}

\begin{figure*}[!tb]
    \centering
    \includegraphics[width=\linewidth]{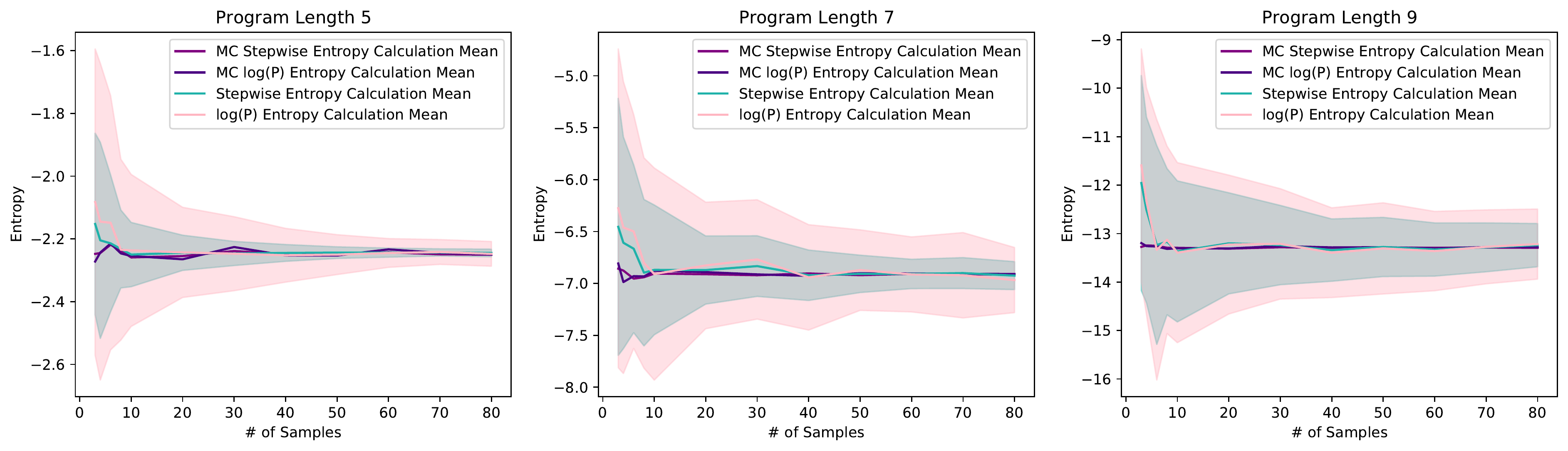}
    \caption{ Compare the entropy estimation following the Equation \ref{eq: BS beam search entropy} as a weighted sum of stepwise entropy versus taking the average of the sequence log probability. The number of samples varies from 2 to 80 on the $x$-axis. The shaded area represents the standard deviation of each estimator. From left to right, we demonstrate the result on datasets of three program lengths. }
    \label{fig:entropy_percentile}
\end{figure*}

\subsubsection{Comparison With Supervised Pretraining}\label{2d cad comparison}

In this section, we measure the image similarity directly in Chamfer Distance (CD) (Equation \ref{eq: schamfer}) for comparison. We pretrained a model on 300k, 150k and 30k ground truth programs (including duplicates) each with learning rate of 0.001. We selected the pretrained models that reaches the lowest CD (at 1.41, 2.00, 2.79 respectively) on a synthetic validation set. We further fine-tuned the pretrained models on the CAD dataset with learning rate 0.006. The transition between pretraining and fine-tuning is delicate. The grammar structure of the output programs (Figure \ref{fig:supervised_pretrain_RL}) collapses when we set the learning rate to be 0.01 (as opposed to 0.006) or fine-tune an unconverged pretrain model. Our model was trained directly on the CAD dataset without supervision with learning rate of 0.01 and entropy coefficient (ec) of 0.009. Entropy coefficient trades off between exploitation and exploration. Higher ec leads to slower convergence but the model is less likely to be stuck at local optimum. Setting ec in the range between 0.005 and 0.012 for this problem has not impact the result significantly. We report the result of the three pretrained models, vanilla RL fine-tuned models, SWOR RL fine-tuned models  as well as our model after beam search with $k = 1, 3, 5$ in Table \ref{table: cad}. 

\begin{table}[]
    \centering
    \begin{tabular}{||c c c c||}
     
     \hline
      Chamfer Distance $\downarrow$ & k = 1 & k =3 & k = 5  \\ 
     \hline\hline
     30k Supervised Model  & 4.09 & 3.38 & 3.02 \\ 
     30k RL Refined Model & 1.92 & 1.83 & 1.79 \\ 
     30k SWOR RL Refined Model & 1.96 & 1.82 & 1.77 \\ 
     \hline
     150k Supervised Model  & 3.64 & 2.89 & 2.63 \\ 
     150k RL Refined Model & 1.91 & 1.79 & 1.73 \\ 
     150k SWOR RL Refined Model & 1.93 & 1.79 & 1.73 \\ 
     \hline
     300k Supervised Model  & 3.32 & 2.69 & 2.38 \\ 
     300k RL Refined Model & 1.66 & 1.54 & 1.50 \\ 
     300k SWOR RL Refined Model & 1.65 & 1.53 & 1.49 \\ 
     \hline
     Unsupervised Model & 1.51 & 1.48 & \textbf{1.47} \\ 
     \hline
   \end{tabular}
    \caption{Empirical comparison of supervised model pretrained on 30k, 150k and 300k programs, their fine-tuned models and our model on 2D CAD dataset with Chamfer distance}
    \label{table: cad}
\end{table}

The poor performance of the pretrained model shows that it is not able to directly generalize to the novel dataset due to the mismatch of the training dataset and the CAD dataset. The unsupervised method exceeds the performance of a supervised model by a large margin because it \emph{was trained directly on the target domain}. It also removes the hyper-parameter tuning step in transition to the RL fine-tuning and reaches a result competitive to a refined model. The unsupervised method performs better when $k=1, 3$ than the fine-tuned models. At convergence, the results of the fine-tuned model pretrained on 300k synthetic data and the unsupervised model become very close at $k = 5$. The two types of refined models reach similar result given same amount of pretraining synthetic data -- poorer performance on less data -- confirming that the quality of the pretraining dataset is a limiting factor for the downstream models.


\subsection{Variance Study of Entropy Estimation}

This study (Figure \ref{fig:entropy_percentile}) investigates the empirical relationship between the two variance estimators and verifies Lemma \ref{entropy lemma} that $\hat{\mathcal{H}}_D$ achieves lower variance than $\hat{\mathcal{H}}$ (Section \ref{exploration}).

We take a single model saved at epoch 40 during the training time of the length 5, 7, and 9 dataset and estimate the entropy with $\hat{\mathcal{H}}_D$ (Equation \ref{eq: stepwise entropy}) and $\hat{\mathcal{H}}$ (Equation \ref{eq:naive entropy}). We consider two sampling schemes: with and without replacement. We combine both entropy estimation methods with the two sampling schemes creating four instances for comparisons. The $x$-axis of the plot documents the number of samples to obtain a single estimation of the entropy.
We further repeat the estimation 100 times to get the mean and variance. The means of SWR method act as a baseline for the means of the SWOR while we compare the standard deviations (the shaded area) of the two entropy estimation methods. 

Across all three datasets, $\hat{\mathcal{H}}_D$ (green) shows significantly smaller variance with the number of samples ranges from 2 to 80. But we notice that longer programs, or more complex images, require much more samples to reduce variance. This makes sense because the search space increases exponentially with the program length. The initial bias in the SWOR estimation dissipates after the number of samples grows over 10 and is greater in dataset with longer program length.

\section{Discussion}

Current program synthesis approaches for non-differentiable renderers employ a two-step scheme which requires the user to first generate a synthetic dataset for pretraining and then use RL fine-tuning with target images. A purely RL-driven approach does not require the curation of a pretraining dataset and can learn directly from target images. Further, unlike approaches that rely on supervised pretraining, the RL approach does not restrict the model to only the program(s) in the synthetic dataset when multiple equal reconstructions exist. This limitation of the pretraining dataset can further impact the downstream models' ability to generalize. In this paper we introduced the first unsupervised algorithm capable of parsing CSG images---created by a non-differentiable renderer---into CFG programs without pretraining. We do so by combining three key ingredients to improve the sample efficiency of a REINFORCE-based algorithm: (i) a grammar-encoded tree LSTM to constrain the search space; (ii) entropy regularization for trading off exploration and exploitation; (iii) sampling without replacement from the CFG syntax tree for better estimation. Our ablation study emphasizes the qualitive and quantitative contributions of each design component. Even though our RL approach does not have access to a pretraining dataset it achieves stronger performance than a supervised method on the synthetic 2D CSG dataset. It further outperforms the supervised pretrained model on the novel 2D CAD dataset and is competitive to the RL fine-tuned model.

\newpage

\bibliography{zhou_171}

\newpage
\appendix

\section{Grammar Tree LSTM Guide}
\label{grammar tree lstm tracing}
This section provides a guide for the tree-LSTM illustration in Figure \ref{fig:tree_lstm}. This guide follows the arrows in the illustration (Figure \ref{fig:tree_lstm}) from left to right:
\begin{itemize}
    \item A grammar stack with a start token $S$ and an end token $\$$ as well as an empty image stack is initialized.
    \item In the first iteration, the token $S$ is popped out. Following Rule \eqref{rule1}, all other options will be masked except $E$, the only possible output. $E$ token is added to the stack.
    \item In the second iteration, or any iteration where the token $E$ is popped, the input for all examples and all softmax outputs are masked except the entries representing $EET$ and $P$ according to Rule \eqref{rule2}. If $EET$ is sampled,  $T$, $E$ and $E$ tokens will be added to the stack separately in that exact order to expand the program further. If $P$ is sampled, it will be added to the stack and the program cannot expand further. 
    \item If $T$ is popped out of the stack, the output space for that iteration will be limited to all the operations (Rule \eqref{rule3}). Similarly, if $P$  is popped out, the output space is limited to all the geometric shapes (Rule \eqref{rule4}).
    \item  When a shape token is sampled, it will not be added to the grammar stack as they do not contribute to the program structure. Instead, the image of the shape will be pushed onto the corresponding image stack.
    \item When an operation token is sampled, it also will not be added to the grammar stack. Instead, we pop out the top two images to apply the operation on them and push the final image onto the image stack again.
    \item When the stack has popped out all the added tokens, the end token $\$$ will be popped out in the last iteration. We then finish the sampling as standard RNN language models.  
\end{itemize}
In practice, we implement the masking mechanism by adding a vector to the output before passing into logsoftmax layer to get the probability. 
The vector contains $0$ for valid output and large negative numbers for invalid ones. This makes sure that invalid options will have almost zero probability of being sampled. The input of the RNN cell includes encoded target image and intermediate images from the image stack, embedded pop-out token from grammar stack and the hidden state from the RNN's last iteration. The exact algorithm is in Algorithm \ref{alg:tree lstm}.

\section{Sampling Without Replacement} \label{sampling without replacement}

This section describes how we achieve sampling without replacement with the help of stochastic beam search \citep{kool2019stochastic}.

At each step of generation, the algorithm chooses the top-$k+1$ beams to expand based on the $\vec{\mathbf{G}}_{\bm{\phi}_{i, j}}$ score at time step $j$ in order to find the top-$k$ stochastic sequences at the end. The use for the additional beam will be explained later. Let $A$ be all possible actions at time step $j$, $\vec{\bm{\phi}}_{i, j} \in \mathbb{R}^A$ is the log probability of each outcomes of sequence $i$ at time $j$ plus the log probability of the previous $j-1$ actions. 

\begin{align}
     \vec{\bm{\phi}}_{i, j} =& \big [\log P(a_1), \log P(a_2), \dots, \log P(a_A) \big] + \notag \\
    &\log P(a_{t_1}, a_{t_2}, \cdots, a_{t_{j-1}}) \cdot \vec{\mathbbm{1}} \notag\\
    =&\big [\log P(a_1), \log P(a_2), \dots, \log P(a_A) \big] +  \phi_{i, j-1} \cdot \vec{\mathbbm{1}}
\end{align}

For each beam, we sample a Gumbel random variable $G_{\phi_{i, j, a}} = \text{Gumbel}(\phi_{i, j, a})$ for each of the element $a$ of the vector $\vec{\bm{\phi}}_{i, j}$.  Then we need to adjust the Gumbel random variable by conditioning on its parent's adjusted stochastic score $G_{s^i_{j-1}}$ being the largest (Equation \ref{eq: G prime}) in relation to all the descendant elements in $\vec{\mathbf{G}}_{\phi_{i, j}}$, the resulting value $\vec{\tilde{\mathbf{G}}}_{\phi_{{i, j}}} \in \mathbb{R}^A$ is the adjusted stochastic score for each of the potential expansions.

\begin{align} \label{eq: G prime}
     \vec{\Tilde{\mathbf{G}}}_{\phi_{{i, j}}}=  &-\log(\exp{(- \vec{\mathbbm{1}} \cdot G_{s^i_{j-1}})} - \exp{(- \vec{\mathbbm{1}} \cdot Z_{i, j})} + \notag\\ &\exp{(-\vec{\mathbf{G}}_{\boldsymbol{\phi}_{{i, j}}}))}
\end{align}

Here $Z_{i, j}$ is the largest value in the vector $\vec{\mathbf{G}}_{\boldsymbol{\phi}_{{i, j}}} = [G_{\phi_{i, j, a_1}}, G_{\phi_{i, j, a_2}}, \cdots, G_{\phi_{i, j, a_A}}]$ and $ G_{s^i_{j-1}}$ is the adjusted stochastic score of $i$-th beam at step $j-1$ from the last iteration. Conditioning on the parent stochastic score being the largest in this top-down sampling scheme makes sure that each leaf's final stochastic score $G_{s^i} $ is independent, equivalent to sampling the sequences bottom up without replacement \citep{kool2019stochastic}. 

Once we have aggregated all the adjusted stochastic scores in $\vec{\Tilde{\mathbf{G}}}_{\phi_{{i, j}}}$ from all previous $k+1$ beams, we select the top-$k+1$ scored beams from $(k+1) * A$ scores for expansions. The selected $k+1$ adjusted stochastic scores become the $G_{s^i_j} \; \forall i$ in the new iteration. Note that the reason that we maintain one more beam than we intended to expand because we need the $k+1$ largest stochastic score to be the threshold during estimation of the entropy and REINFORCE objective. This is explained next. Please refer to Algorithm \ref{alg:beam search} for details in the branching process.

The sampling without replacement algorithm requires importance weighting of the objective functions to ensure unbiasness. The weighting term is $\frac{p_\theta (s^i)}{q_{\theta, \kappa}(s^i)}$. $p_{\theta}(s^i)$ represents the probability of the sequence $s^i$ ($s^i$ is the $i$-th completed sequence and $p_\theta (s^i) = \exp{\phi_i}$) and $S$ represents the set of all sampled sequences $s^i$ for $i = 1, 2, \cdots, k$. $q_{\theta, \kappa}(s^i_j) = P(G_{s^i_j} > \kappa) = 1 - \exp (- \exp (\phi_{i, j} - \kappa))$, where $\kappa$ is the ($k+1$)-th largest $G_{s^i_j}$ score for all $i$ and $\phi_{i, j} = \log P(a_{t_1}, a_{t_2}, \cdots, a_{t_{j}})$ is the log likelihood of the first $j$ actions of $i$-th sequence, can be calculated based on the CDF of the Gumbel distribution. It acts as a threshold for branching selection. We can use the log probability of the sequence here to calculate the CDF because the adjust stochastic scores $G_{s^i_j}$'s are equivalent to the Gumbel scores of sequences sampled without replacement from bottom up. During implementation, we need to keep an extra beam, thus $k+1$ beams in total, to accurately estimate $\kappa$ in order to ensure the unbiasness of the estimator. 

To reduce variance of our objective function, we introduce additional normalization terms as well as a baseline. However, the objective function is biased with these terms. The normalization terms are $W(S) = \sum_{s^i \in S} \frac{p_\theta (s^i)}{q_{\theta, \kappa} (s^i)} $ and $W^i(S) = W(S) - \frac{p_\theta (s^i)}{q_{\theta, \kappa}(s^i)} + p_\theta (s^i)$.

Incorporating a baseline into the REINFORCE objective is a standard practice.  A baseline term is defined as $B(S) = \sum_{s^i \in S} \frac{p_\theta (s^i)}{q_{\theta, \kappa} (s^i)} f(s^i)$ and $f(s^i)$ should be the reward of the complete $i$-th program $s^i$ in this case.

To put everything together, the exact objective is as follows \citep{kool2019buy}:
\begin{equation}\label{eq: BS reinforce}
\nabla_\theta \mathbb{E}_{s \sim p_\theta (s)} [ f(s) ] \approx \sum_{s^i \in S} \frac{1}{W^i (S)} \cdot \frac{\nabla_\theta p_\theta (s^i)}{q_{\theta, \kappa} (s^i_n)} (f(s^i) - \frac{B(S)}{W(S)})
\end{equation} 
Entropy estimation uses a similar scaling scheme as the REINFORCE objective:
\begin{align} \label{eq: BS beam search entropy}
    &\hat{\mathcal{H}}_{D}(X_1, X_2, X_3, \cdots, X_n) \approx \notag\\
    &\sum^n_{j=1} \frac{1}{W_j(S)} \sum_{s^i \in S} \frac{p_\theta (s^i_j)}{q_{\theta, \kappa} (s^i_j)} \mathcal{H}(X_j | X_1 = x^i_1, \cdots, 
     X_{j-1} = x^i_{j-1})
\end{align}
where $W_j(S) = \sum_{s^i \in S} \frac{p_\theta (s^i_j)}{q_{\theta, \kappa} (s^i_j)}$ and $s^i_j$ denotes the first $j$ elements of the sequence $s^i$. The estimator is unbiased excluding the $\frac{1}{W_j(S)}$ term.

\section{Proof of Stepwise Entropy Estimation's Unbiasness} \label{unbiasness}

Entropy of a sequence can be decomposed into the sum of the conditional entropy at each step conditioned on the previous values. This is also called the chain rule for entropy calculation. Let $X_1, X_2, \cdots, X_n$ be drawn from $P(X_1, X_2, \cdots, X_n)$ \citep{cover2012elements}:
\begin{align}
    \mathcal{H}(X_1, X_2, \cdots, X_n) = \sum^n_{j = 1} \mathcal{H}(X_j| X_1, \cdots, X_{j-1}) 
\end{align}
If we sum up the empirical entropy at each step after the softmax output, we can obtain an unbiased estimator of the entropy. Let $S$ be the set of sequences that we sampled and each sampled sequence $s^i$ consists of $X_1, X_2, \cdots, X_n$:
\begin{align*}
    &\mathbb{E}_{X_1, \dots, X_{j-1}}(\hat{\mathcal{H}}_D) \notag \\ 
    &= \mathbb{E}_{X_1, \dots, X_{j-1}} (\frac{1}{|S|} \sum_{i \in |S|} \sum^n_{j = 1} \mathcal{H}(X_j | X_1 = x^i_1 , \dots, \\
    & \quad X_{j-1} = x^i_{j-1}))\\
    &= \frac{1}{|S|} \cdot |S| \sum^{n}_{j=1} \mathcal{H}(X_j | X_1, \cdots, \quad X_{j-1}) \\
    &= \mathcal{H}(X_1, X_2, \cdots, X_n) 
\end{align*}
In order to incorporate the stepwise estimation of the entropy into the beam search, we use the similar reweighting scheme as the REINFORCE objective. The difference is that the REINFORCE objective is reweighted after obtaining the full sequence because we only receive the reward at the end and here we reweight the entropy at each step.  We denote each time step by $j$ and each sequence by $i$, the set of sequences selected at time step $j$ is $S_j$ and the complete set of all possible sequences of length $j$ is $T_j$ and $S_j \in T_j$. We are taking the expectation of the estimator over the $G_{\phi_{i, j}}$ scores. As we discussed before, at each step, each potential beam receives a stochastic score $G_{\phi_{i, j}}$. The beams associated with the top-$k+1$ stochastic scores are chosen to be expanded further and $\kappa$ is the $k+1$-th largest $G_{\phi_{i, j}}$. $\kappa$ can also be seen as a threshold in the branching selection process and $q_{\theta, \kappa}(s_j^i) =P(G_{s^i_j} > \kappa)= 1 - \exp (- \exp (\phi_{i, j} - \kappa))$. For details on the numerical stable implementation of $q_{\theta, \kappa}(s_j^i) $, please refer to \citep{kool2019stochastic}.
\begin{align*}
    &\mathbb{E}_{G_{\phi}}(\sum^n_{j=1} \sum_{s^i_j \in S_j} \frac{p_\theta (s^i_j)}{q_{\theta, \kappa} (s^i_j)} \mathcal{H}(X_j | X_1 = x^i_1, X_2=x^i_2, \cdots, \notag\\ 
    & \quad X_{j-1} = x^i_{j-1})) \notag\\
    &=\sum^n_{j=1} \mathbb{E}_{G_{\phi}}(  \sum_{i \in |T_j|} \frac{p_\theta (s^i_j)}{q_{\theta, \kappa} (s^i_j)} \mathcal{H}(X_j | X_1 = x^i_1, X_2=x^i_2, \notag \\
    &\quad \cdots, X_{j-1} = x^i_{j-1}))\mathbbm{1}_{\{x^i_1, \cdots, x^i_j\} \in S_j}) \notag \\
    &= \sum^n_{j=1} \sum_{i \in |T_j|} p_\theta (s^i_j)  \mathcal{H}(X_j | X_1 = x^i_1, X_2=x^i_2, \cdots, \notag\\
    &\quad X_{j-1} = x^i_{j-1}) \mathbb{E}_{G_{\phi}} (\frac{ \mathbbm{1}_{\{s^i_j = x^i_1, \cdots, x^i_j\} \in S_j}}{q_{\theta, \kappa} (s^i_j)}) \\
    &=  \sum^n_{j=1}  \mathcal{H}(X_j | X_1 , X_2, \cdots, X_{j-1}) \cdot 1 \notag\\
    &= \mathcal{H}(X_1, X_2, \cdots, X_n) 
\end{align*}
For the proof of $\mathbb{E}_{G_{\phi}} (\frac{ \mathbbm{1}_{\{s^i_j \in S_j}}{q_{\theta, \kappa} (s^i_j)}) = 1$, please refer to \citep{kool2019stochastic}, appendix D.

\section{Proof of Lower Variance of the Stepwise Entropy Estimator}\label{lowvariance}
   
We will continue using the notations from above. We want to compare the variance of the two entropy estimator $\mathcal{\hat{H}}$ and the stepwise entropy estimator $\mathcal{\hat{H}}_D$ and show that the second estimator has lower variance. 
\begin{proof}
We abuse  $\mathbb{E}_{X_j}$ to be $\mathbb{E}_{X_j|X_1,\dots,X_{j-1}}$ and $\mathrm{Var}_{X_j}$ to be $\mathrm{Var}_{X_j|X_1,\dots,X_{j-1}}$ to simplify the notations.
\begin{align*}
    &\mathrm{Var}_{X_1, X_2, \cdots, X_n} (\frac{1}{|S|}\sum_{i \in |S|} \sum^n_{j = 1} \mathcal{H}(X_j | X_1 = x^i_1 , \dots, \\
    & \quad X_{j-1} = x^i_{j-1})) \\
    &= \frac{1}{|S|^2} \sum_{i \in |S|} \sum^n_{j = 1}  \mathrm{Var}_{X_1, X_2, \cdots, X_n}(  \mathcal{H}(X_j | X_1 = x^i_1 , \dots, \\
    & \quad X_{j-1} = x^i_{j-1})) \\ 
    &= \frac{1}{|S|^2} \sum_{i \in |S|} \sum^n_{j = 1} ( \mathbb{E}(  \mathcal{H}^2(X_j | X_1 = x^i_1 , \dots, X_{j-1} = x^i_{j-1})) \\
    & \quad -   \mathbb{E}^2(  \mathcal{H}(X_j | X_1 = x^i_1 , \dots, X_{j-1} = x^i_{j-1}))  )\\ 
    & = \frac{1}{|S|^2} \sum_{i \in |S|} \sum^n_{j = 1}  ( \mathbb{E}_{X_1, \cdots, X_{j-1}}  \mathbb{E}^2_{X_j}( \log P(X_j | X_1 = x^i_1 , \\
    & \quad \dots, X_{j-1} = x^i_{j-1})) \\
    & \quad -  \mathbb{E}^2_{X_1, \cdots, X_{j-1}}  \mathbb{E}_{X_j}( \log P(X_j | X_1 = x^i_1 , \dots, \\
    & \quad X_{j-1} = x^i_{j-1}))  ) \\ 
    & = \frac{1}{|S|^2} \sum_{s^i \in S} \sum^n_{j = 1} ( \mathbb{E}_{X_1, \cdots, X_{j-1}} (\mathbb{E}_{X_j}( \log^2 P(X_j | X_1 = x^i_1 , \\
    &\quad   \dots, X_{j-1} = x^i_{j-1}))  \\
    & \quad - \mathrm{Var}_{X_j} (\log P(X_j | X_1 = x^i_1 , \dots, X_{j-1} = x^i_{j-1})) )  \\
    & \quad - \mathbb{E}^2_{X_1, \cdots, X_{j-1}}  \mathbb{E}_{X_j}( \log P(X_j | X_1 = x^i_1 , \dots,  \\
    & \quad X_{j-1} = x^i_{j-1}))  ) \\ 
    & = \frac{1}{|S|^2} \sum_{i \in |S|} \sum^n_{j = 1} ( \mathbb{E}_{X_1, \cdots, X_{j}}( \log^2 P(X_j | X_1 = x^i_1 , \dots,  \\
    & \quad X_{j-1} = x^i_{j-1})) \\
    & \quad - \mathbb{E}^2_{X_1, \cdots, X_{j}}  ( \log P(X_j | X_1 = x^i_1 , \dots, X_{j-1} = x^i_{j-1}))  )  \\
    &\quad - \mathbb{E}_{X_1, \cdots, X_{j-1}} \mathrm{Var}_{X_j} (\log P(X_j | X_1 = x^i_1 , \dots, \\
    & \quad X_{j-1} = x^i_{j-1}))\\
    & =  \frac{1}{|S|^2} \sum_{ i \in |S|} \sum^n_{j = 1} ( \mathrm{Var}_{X_1, \cdots, X_{j}}( \log P(X_j | X_1 = x^i_1 , \dots, \\
    & \quad X_{j-1} = x^i_{j-1}))\\
    &\quad - \mathbb{E}_{X_1, \cdots, X_{j-1}} \mathrm{Var}_{X_j} (\log P(X_j | X_1 = x^i_1 , \dots,  \\
    & \quad X_{j-1} = x^i_{j-1})) \\
    &\leq \frac{1}{|S|^2} \sum_{i \in |S|} \sum^n_{j = 1} \mathrm{Var}_{X_1, \cdots, X_{j}}( \log P(X_j | X_1 = x^i_1 , \dots, \\
    & \quad X_{j-1} = x^i_{j-1})) \\
    & = \mathrm{Var}_{X_1, X_2, \cdots, X_n} (\frac{1}{|S|}\sum_{i \in |S|} \log P(s^i)) 
\end{align*} 
\end{proof}

The fifth equation holds from the fact that $\mathbb{E}_{X}^2\mathbb{E}_{Y|X}[f(X,Y)] = \mathbb{E}^2_{X,Y}[f(X,Y)]$.
The result still stands after applying reweighting for the beam search.
\begin{figure*}
    \centering
    \includegraphics[width=0.8\linewidth]{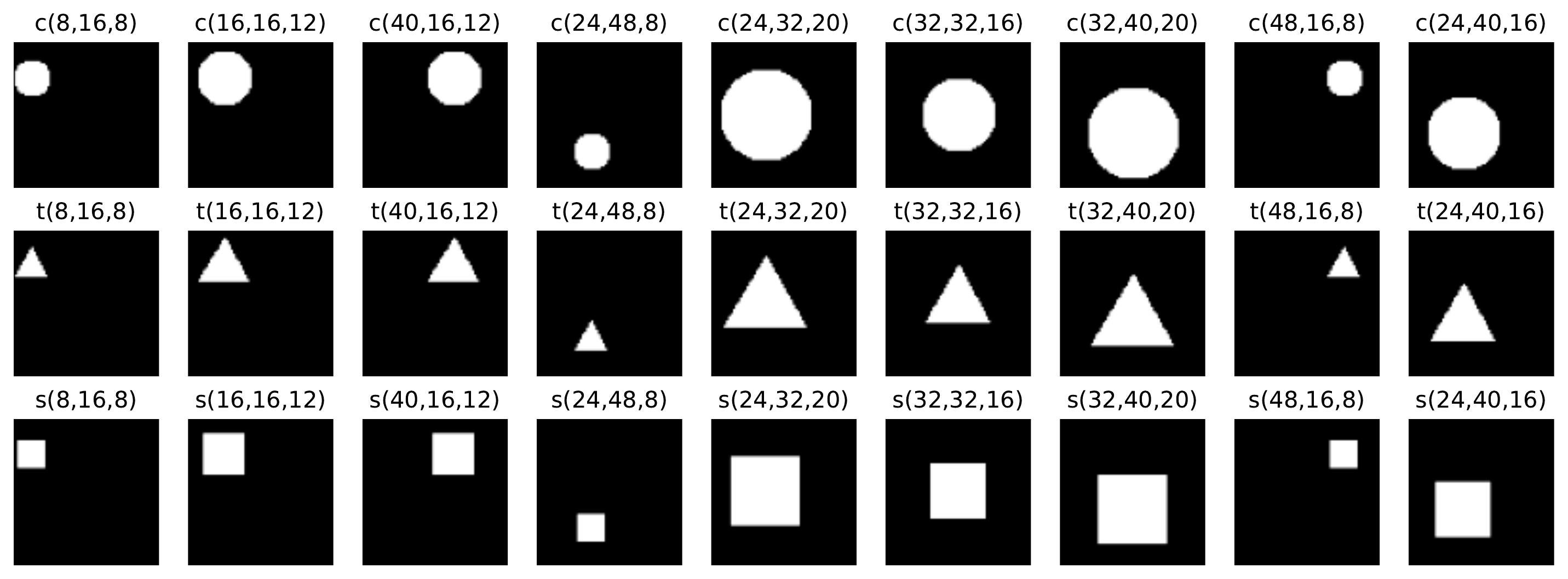}
    \vspace{-10pt}
    \caption{ Each shape encoding is on top of the image it represents.  }
    \vspace{-10pt}
    \label{fig:shape_encoding}
\end{figure*}

\section{Shape Encoding Demonstration}
In Figure \ref{fig:shape_encoding}, we show the code name on top of the image that it represents. c, s, and t represent circle, square and triangle respectively. The first two numbers represent the position of the shape in the image and the last number represents the size. 

\section{Additional Test Output Examples for the 2D CAD Dataset}

\begin{figure*}
    \centering
    \includegraphics[width=\linewidth]{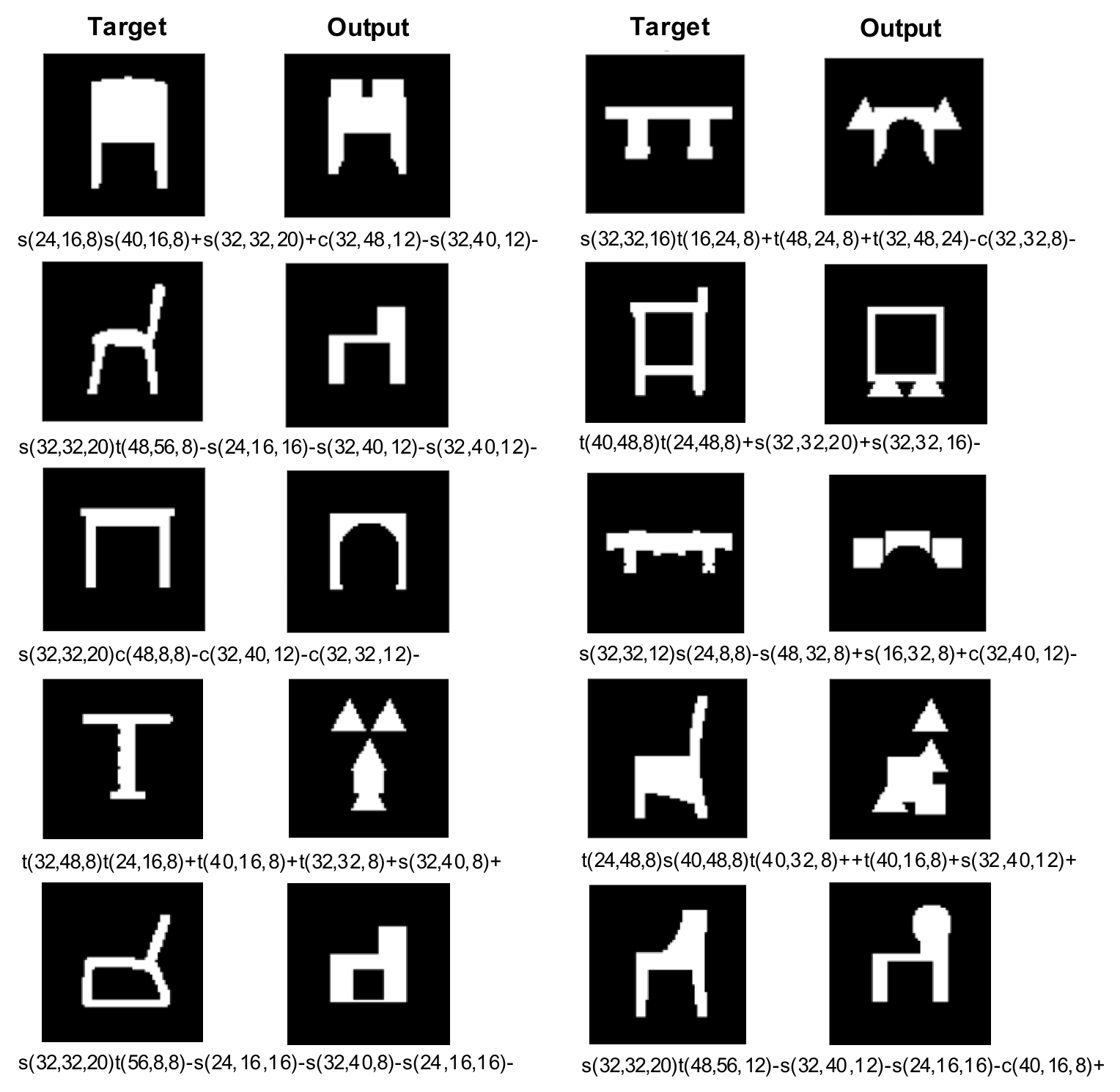}
    \vspace{-10pt}
    \caption{ Additional test output with corresponding programs. The odd-numbered columns contain the target images and the images to their right are example outputs. }
    \vspace{-10pt}
    \label{fig:additional}
\end{figure*}

In this section, we include additional enlarged test outputs (Figure \ref{fig:additional}). We add the corresponding output program below each target/output pair. We observe that the algorithm approximates thin lines with triangles in some cases. Our hypothesis for the cause is the Chamfer Distance reward function being a greedy algorithm and it finds the matching distance based on the nearest features. 

\end{document}